\documentclass{article}

\usepackage[dblblindworkshop,final]{neurips_2026}
\usepackage[T1]{fontenc}
\usepackage[utf8]{inputenc}
\usepackage{booktabs}
\usepackage{microtype}
\usepackage{amsmath}
\usepackage{amssymb}
\usepackage[hidelinks]{hyperref}
\usepackage{tikz}
\usetikzlibrary{arrows.meta,positioning}

\title{Recoverability and Verifier Strength: Two Measured\\
Preconditions for Unattended Agent Self-Improvement}

  \workshoptitle{AgenticOS: Co-designing Systems and ML Foundations of an OS Layer for Agentic AI}

\author{%
  De Wang \\ Uniphore \\ \texttt{de.wang@uniphore.com} \And
  Peng Qi \\ Uniphore \\ \texttt{peng.qi@uniphore.com}%
}
\date{}

\begin{document}
\maketitle

\begin{abstract}
\noindent
Agent platforms are starting to improve their own agents: they read a deployed agent's execution traces,
derive a change, have an automated verifier approve it, and release it with no human
author. How much of this loop can safely run without a human? We propose two
quantities a platform can measure. \emph{Verifier strength} $\sigma$ is how well the available verifier
--- often an LLM judge --- agrees with a trusted reference verifier; it limits how much a step can improve the
agent. \emph{Recoverability} is how cheaply a bad change can be undone: whether undoing it needs a new
verifier approval, and how much of the old behavior a rollback restores. A step should run unattended only if
both are adequate. On our platform this rule allows prompt, tool and skill edits to run unattended and keeps
weight updates under human sign-off.

Our main contribution is these two definitions and how to measure them. We built a platform that applies the
rule and tested it in two ways: on the platform's own approval gate, and in controlled experiments on the
$\tau^2$-bench agent benchmark. On verifier strength, the platform's LLM judge sometimes gave the same score no
matter what it was shown (9 of 37 settings) and reported no error, so a verifier can look healthy while
approving anything; its strength must be measured, and two simple tests catch the failure. On
recoverability, restoring a saved copy brought the agent back to its original behavior, but a regenerated
undo did not, so only changes whose old version is kept can be reversed without a human; and while a change
was deployed, it altered the outcome of $17\%$ of tasks beyond normal run-to-run variation, which no undo repairs.
Together these support the rule: a change can run unattended only when its old version is kept and its
verifier's strength has been measured.
\end{abstract}

\section{Introduction}

Agents built on language models now hold conversations, call tools, and write to business
systems~\citep{yao2024taubench,barres2025tau2}. They also fail often. On the $\tau^2$-bench tasks we study
the baseline agent solves just over half of its tasks, through ordinary mistakes: a wrong tool call, a
missing confirmation, steps taken out of order~\citep{cuadron2026saber}. Fixing such failures means reading
traces, finding the cause, revising the agent, checking the revision and deploying it. Done by hand, this
does not scale with the number of agents a platform hosts, so platforms have begun to automate
it~\citep{wang2026linkedin,li2026audit}.

We call one pass of this automated repair a \emph{self-improvement step}: the platform reads its own
agents' traces, derives a change, checks it with a verifier and deploys it, with no person writing the
change. Inference-time self-correction~\citep{shinn2023reflexion,madaan2023selfrefine} and human-written
fixes do not count. Training and serving became platform services long
ago~\citep{baylor2017tfx,moritz2018ray}; this loop has not, yet it needs the same things: a budget,
controlled access to the deployed agent, a point where changes are admitted, and an audit trail
(\S\ref{sec:abstractions}).

\begin{figure}[t]
\centering
\resizebox{\linewidth}{!}{%
\begin{tikzpicture}[font=\small, >=Stealth, node distance=7mm,
  box/.style={draw, rounded corners, align=center, minimum height=9mm, inner sep=4pt}]
\node[box] (tr) {Agent $A$\\[-1pt]execution traces};
\node[box, right=of tr] (pr) {Propose\\[-1pt]change $\iota$};
\node[box, right=of pr, fill=gray!15] (gt) {Verifier gate\\[-1pt]\textbf{strength} $\sigma$};
\node[box, right=12mm of gt] (lv) {$A'$ serves users\\[-1pt]blast radius $\beta$};
\node[box, right=12mm of lv, fill=gray!15] (rb) {Roll back to $A$\\[-1pt]\textbf{re-approval} $g$, fidelity $\rho$};
\draw[->] (tr) -- (pr);
\draw[->] (pr) -- (gt);
\draw[->] (gt) -- node[above, font=\footnotesize] {admit} (lv);
\draw[->] (lv) -- node[above, font=\footnotesize] {if bad} (rb);
\draw[->] (gt.south) -- ++(0,-5mm) -| node[pos=0.25, below, font=\footnotesize] {reject: keep $A$} (tr.south);
\draw[->, dashed] (rb.north) -- ++(0,5mm) -| node[pos=0.25, above, font=\footnotesize]
  {old value not kept ($g{=}1$): the undo must pass the gate again} (gt.north);
\end{tikzpicture}}
\caption{One self-improvement step. Verifier strength $\sigma$ governs the admit decision; recoverability
governs the cost when an admitted change is bad. If the old value was kept, rollback is a plain write
($g{=}0$); otherwise the undo must be regenerated and approved by the same gate ($g{=}1$).}
\label{fig:loop}
\end{figure}
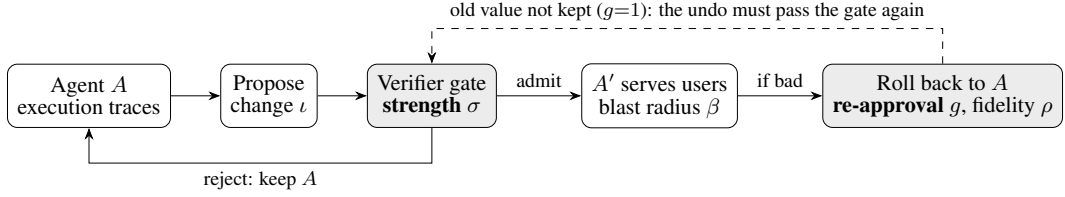

This paper asks how much of the loop may run without a human. Our answer rests on two measurable
quantities (\S\ref{sec:defs}). \textbf{Verifier strength} measures how far the verifier's approvals can be
trusted: its agreement with a reference verifier on a
scale from $-1$ to $1$, where $1$ is perfect agreement and a verifier that always gives the same answer
scores $0$.
\textbf{Recoverability} measures how cheaply a bad change is undone. It records whether undoing a change needs a
new verifier approval and how much of the old behavior actually returns. If no copy of the old version was kept (for example, after fine-tuning a model in place), the undo must pass
the same verifier that just approved the mistake. Each quantity
guards against a different failure, so a loop can be only as autonomous as the weaker one allows
(Figure~\ref{fig:loop}). We derive this rule (\S\ref{sec:governance}), build a platform that enforces it
(\S\ref{sec:abstractions}), and test the assumptions it rests on (\S\ref{sec:results}), first in a case study of our
platform's own admission gate and then in controlled experiments on $\tau^2$-bench. We do not address
adversarial or misaligned agents, only the ordinary failure of admitting a change that should have been
rejected.

\paragraph{Contributions.} Our main contribution is a method: the two definitions, how to measure them, and
the rule they support. (i)~Two measurable preconditions for unattended changes, and a rule that refuses
whenever either is unmeasured (\S\ref{sec:defs}, \S\ref{sec:governance}). (ii)~A platform that enforces the
rule by typing changes by recoverability, declaring verifiers as resources, and recording refusal as its own
outcome (\S\ref{sec:abstractions}). (iii)~Tests of the rule's assumptions: a case study of \emph{silent gate
collapse} in our platform's own admission gate, with two tests that catch it (\S\ref{sec:admission}), and
experiments that vary each quantity separately for prompt edits on two $\tau^2$-bench domains
(\S\ref{sec:tau}). Our verdicts for tool, skill and weight changes follow from the definitions and are not
separately tested.

\section{Definitions}
\label{sec:defs}

We write an agent as $A=(\theta,p,T,K)$: model weights, system prompt, tool schemas and skills.

\paragraph{Self-improvement step.} A change $A\mapsto A'$ that the platform derives from $A$'s own traces,
in the spirit of self-taught training~\citep{zelikman2022star}, and admits through a verifier, with no
human writing it. \emph{Running example:} airline traces show the agent canceling bookings without asking
the user; the loop adds ``confirm before any cancellation'' to $p$, an LLM judge approves the edit, and it
goes live. We measure a step by $\Delta_k$, the change in the share of held-out tasks solved in \emph{all}
$k$ repeated trials, because a platform cares about reliability rather than best cases.

\paragraph{Recoverability.} For a change $\iota$ and its undo $\iota^{-1}$ we measure three things.
\emph{Re-approval} $g(\iota)\in\{0,1\}$ records whether the way back from $A'$ to $A$ needs a new verifier
approval. If the old value was kept, the undo is a plain write and $g{=}0$; this holds for prompt and schema
edits and for a retained model checkpoint. If no old value was kept, as after fine-tuning from the current
checkpoint, the undo must itself be generated and approved, so $g{=}1$. Both other
measures use a fixed set of probe tasks $P$, scored pass or fail. \emph{Restoration fidelity} $\rho(\iota)$
asks whether the undo brings the old behavior back: it is the share of tasks in $P$ with the same outcome
after the change is undone as before the change was made. \emph{Blast radius} $\beta(\iota)$ is the share
of tasks in $P$ whose outcome changed while $\iota$ was deployed, relative to the unchanged agent: what even
a perfect rollback cannot recover, because those outcomes were already delivered. In our experiments (\S\ref{sec:tau}), $P$
is the 20 held-out test tasks for $\rho$ (the 30 training tasks for the regenerated undos) and all 50
airline tasks for $\beta$. The agent and the simulated user are both stochastic, so we also run a \emph{null change}
(no change at all), which gives noise floors $\rho_0<1$ and $\beta_0>0$; only values beyond these floors
count. Because $\beta$ covers only the probe tasks, it is a lower bound: real side effects such as an issued
refund must be tracked at the tool boundary, where recent work locates recoverability for
agents~\citep{yang2026dart,zheng2026acrfence}.

\paragraph{Verifier strength.} A verifier scores a trajectory as pass or fail; it may be a program or an
LLM judge~\citep{zheng2023judging}. Given a trusted reference verifier $v^\star$ and a set of trajectories
$C$, strength is the Matthews correlation $\sigma(v)=\mathrm{MCC}(v,v^\star)\big|_C\in[-1,1]$. We use MCC
rather than accuracy because a verifier that always gives the same answer scores exactly $\sigma{=}0$,
whereas its accuracy equals the share of the more common verdict and can look respectable.

\section{The autonomy rule}
\label{sec:governance}

\paragraph{The rule.} A change may go live without human sign-off only if $g(\iota){=}0$ and
$\sigma(v)\ge\sigma_{\min}$, where $\sigma_{\min}$ reflects the cost of wrongly admitting a bad change. If
either quantity is unmeasured, the platform refuses. Our data do not fix $\sigma_{\min}$: the strength
below which weak verifiers start to hurt moves by $\pm0.15$ with the number of simulations the gate averages, so a
threshold from one benchmark would not transfer (App.~\ref{app:sigmamin} gives the numbers and a
calibration procedure). In the running example the old prompt is kept, so $g{=}0$, and $\sigma$ is the
judge's agreement with $\tau^2$-bench's database check; the edit goes live only if that $\sigma$ was measured
and clears $\sigma_{\min}$. A fine-tune with no checkpoint kept has $g{=}1$ and is refused however good
the judge is.

\paragraph{Why both conditions.} A self-improvement step does two things: it \emph{decides} whether to admit
$A\to A'$, and then it \emph{changes} a live system. Each quantity bounds one of these failures. Verifier
strength bounds the decision, because the verdict is the loop's only evidence that $A'$ is better than $A$;
at $\sigma{=}0$ the verdict says nothing about quality~\citep{zheng2023judging}. Recoverability bounds the
change. At $g{=}0$ the platform writes back the old value and pays only for the traffic served while the
change was deployed, which is what $\beta$ measures. At $g{=}1$ the old value is gone,
so the way back must be regenerated and approved by the same gate that made the mistake. Neither quantity can
make up for the other, so autonomy is limited by the weaker one rather than traded between them. A strong
verifier cannot justify an irreversible change, since its errors would then be permanent, and easy rollback
cannot justify a weak verifier, since the loop would only churn the agent.

\paragraph{Assumptions to test.} The rule rests on three assumptions, which \S\ref{sec:results}
tests: (a)~a verifier's strength cannot be taken for granted and must be measured; (b)~a
weaker verifier admits worse changes; and (c)~keeping the old value makes rollback exact and losing it does not,
while an admitted change still affects tasks that rollback cannot undo.

\begin{figure}[t]
\centering
\begin{minipage}[c]{0.64\linewidth}
\centering
\begin{tikzpicture}[font=\footnotesize,
  cell/.style={draw, align=center, text width=34mm, minimum height=11mm, inner sep=1.5pt},
  ok/.style={cell, fill=gray!25}, no/.style={cell}]
\node[ok] (a) at (0,0) {\textbf{Run unattended}\\prompt, tool-schema\\and skill edits};
\node[no] (b) at (36mm,0) {\textbf{Human sign-off}\\irreversible, even with\\a strong verifier};
\node[no] (c) at (0,-12mm) {\textbf{Human sign-off}\\weight updates;\\data filter (disabled)};
\node[no] (d) at (36mm,-12mm) {\textbf{Human sign-off}\\fine-tune with no\\checkpoint kept};
\node[above=1mm of a, font=\footnotesize\bfseries] {old value kept ($g{=}0$)};
\node[above=1mm of b, font=\footnotesize\bfseries] {old value gone ($g{=}1$)};
\node[left=1mm of a, font=\footnotesize\bfseries, align=right] {$\sigma\ge\sigma_{\min}$,\\measured};
\node[left=1mm of c, font=\footnotesize\bfseries, align=right] {$\sigma$ low or\\unmeasured};
\end{tikzpicture}
\end{minipage}\hfill
\begin{minipage}[c]{0.33\linewidth}
\caption{The rule as a matrix, with our platform's intervention types placed. A change runs unattended
only when both preconditions hold (shaded); every other cell needs a human.}
\label{fig:rule}
\end{minipage}
\end{figure}

\section{The platform: enforcing the rule}
\label{sec:abstractions}

\paragraph{What the platform does.} Our platform improves deployed agents from their own production traces. For each agent it
(1)~collects execution traces and classifies failures, such as a wrong tool sequence, outdated knowledge or
a degraded prompt; (2)~maps each failure to a change along one of four dimensions: rewriting the system
prompt, fixing a tool definition, updating a skill, or fine-tuning a small model on distilled traces;
(3)~builds the candidate and evaluates it; and (4)~promotes it or stops. Each attempt runs as a recorded
session, so every deployed change can be traced back to the evidence that admitted it, as in production ML
pipelines~\citep{baylor2017tfx}. We built this loop to enforce the rule; three design choices put the rule
into it.

\paragraph{Changes are typed by recoverability.} An agent can be changed in four ways --- weights, system
prompt, tool definitions and skills --- and each is a separate resource type that records how it is undone
and hence its $g$ (App.~\ref{app:rev}). This lets ``may this proceed without a human?'' have a
different answer for each type (Figure~\ref{fig:rule}). Prompt, tool-schema and skill edits are undone by a
write and may promote automatically. Weight updates can be undone by re-pinning the previous adapter, but
their verifier is a learned judge of the kind \S\ref{sec:admission} shows can fail silently, so they need
human sign-off. A training-data filter with no workable verifier ships disabled.

\paragraph{Evaluation decides what goes live.} A proposed change is not deployed until evaluation approves
it, so evaluation works like an admission check in other platforms~\citep{k8sadmission}: if it fails or
cannot run, nothing is deployed; every decision is logged; and no caller can skip it. Each agent declares
which verifiers it uses, with each verifier's measured $\sigma$, so before promoting a change the platform can
check whether an adequate verifier exists.

\paragraph{Refusal is a recorded decision.} The rule has three outcomes, not two: admit, reject (the
change is worse), and refuse (the platform cannot decide unattended because $\sigma$ is unmeasured or the
old value is gone). Ours records each refusal with the missing precondition, so a human sees what is waiting and why
instead of a change that silently waits forever.

\section{Experiments and results}
\label{sec:results}

We test the three assumptions in two settings: a case study of our platform's own admission gate
(\S\ref{sec:admission}), and controlled experiments on $\tau^2$-bench (\S\ref{sec:tau}). Table~\ref{tab:summary}
summarizes what each result implies.

\subsection{Case study: silent gate collapse in our platform}
\label{sec:admission}

This case study tests assumption~(a), that a verifier's strength must be measured rather than assumed. Before our platform fine-tunes a
model, it collects agent traces into a training set, and an LLM-judge gate scores the set's diversity; a set
that scores too low is not used for training. This gate is a verifier in the sense of \S\ref{sec:defs}: it
decides whether a change, here a weight update, may proceed. It had been treated as working because it
always returned a score. (A second gate, which filters individual training rows, ships disabled;
App.~\ref{app:design}.)

\paragraph{A diversity gate that ignores half its instructions.}
\label{sec:judge}
The gate's LLM judge~\citep{zheng2023judging} scores the training set from 0 to 1. Its prompt asks for two things: reward \emph{breadth} (many distinct seed tasks) and
penalize \emph{imbalance} (a few seeds dominating); the input gives the imbalance as a number. We tested 3
judge models, 4 prompt variants, 4 imbalance levels, breadths 3--12 and two agent configurations: 172
settings of 10 runs each. In the deployed setting the gate ignores imbalance completely. It returned 0.250 in
all 40 runs with zero spread, exactly what it returns when every mention of imbalance is removed from the
prompt. This is not an artifact of greedy decoding, since 98 of the 172 settings vary. Full collapse,
the same score at all four imbalance levels, occurs in 9 of 37 configurations, all on the deployed judge, the
smallest of the three. That judge also ran most configurations (33 of 37), so the count alone does not
implicate it; a matched comparison does: on identical inputs the deployed judge stays flat while both larger
judges track imbalance ($p<10^{-11}$ each; App.~\ref{app:design}).

\emph{Implication.} We call this \emph{silent gate collapse}: a gate sound in one configuration and empty
in another, with no error to show the difference. A constant score looks like a confident answer; this is the $\sigma{=}0$ point of \S\ref{sec:defs},
and it supports assumption~(a).

Two cheap tests catch this, each costing a handful of calls, and both belong in the platform.
\textbf{Delete the signal:} remove the property the gate claims to measure and check that its score changes.
\textbf{Check the guarantees:} verify that request-level settings the gate relies on actually took effect;
three of ours did not, forced tool calls among them (App.~\ref{app:tests}). What did work was stating the
criterion as a checkable bound: ``if the largest seed exceeds 50\%, score at most 0.10'' held in all 16
settings above 50\%, where the original wording never scored that low (30 of 30 runs vs.\ 0 of 30 on matched
inputs, Fisher $p<10^{-16}$; Table~\ref{tab:judge}).

\subsection{Controlled experiments on $\tau^2$-bench}
\label{sec:tau}

Assumptions~(b) and~(c) need controlled conditions, since on a live platform the two quantities change
together. We fix one and vary the other: experiment A tests~(b), experiment B tests~(c). Both use $\tau^2$-bench, whose agents match ours
(multi-turn, tool-using, database-backed, with a simulated user) and whose task score is the product of a
configurable subset of five checks~\citep{barres2025tau2}. Changing the
subset changes the verifier $v$ while leaving tasks, agent and environment fixed, and scoring is separate from
running, so one set of trajectories can be re-scored under every verifier.

\paragraph{Setup.} We use the airline domain: 50 tasks split 30 for training and 20 for testing, with
natural-language assertions on every task, so a program verifier and an LLM judge score the \emph{same}
trajectories, the comparison \S\ref{sec:admission} could not make in production. The reference verifier combines
database-state equality with communication checks; the latter apply to only 4 of 80 test trajectories, so on
the other 76 the reference is database equality alone. The loop is ours,
reduced to one recipe: read failed training trajectories, propose a system-prompt revision, admit it if the
verifier under test prefers it, and measure the result on the test split with the reference verifier. We
report mean pass rate rather than $\mathrm{pass}^k$, which at four trials discards partial credit.
\textbf{A1} measures $\sigma$ for real verifiers; \textbf{A2} tests whether lower $\sigma$ picks worse
changes; \textbf{B} fixes the verifier at the reference and compares restoring a stored copy ($g{=}0$) with
regenerating the undo ($g{=}1$). Weight updates are not tested. Each agent is run twice per training task and four times per test task. Intervals are
$95\%$ task-resampling bootstrap intervals (App.~\ref{app:pool}), paired for differences~\citep{miller2024errorbars}. Where the
hypothesis fixes a direction we report one-sided $p$-values, and we correct across candidates with Holm's
method.

\paragraph{Real verifiers (A1).} Table~\ref{tab:sigma} gives $\sigma$ for each verifier on one set of 80
test trajectories (20 tasks, 4 trials each). The LLM judge reaches $\sigma{=}0.70$ against the program check
on the \emph{same} trajectories, far above zero (Fisher $p<10^{-9}$) but far from perfect. Its errors are
clustered: all 4 false admits come from one task and the 8 false rejects from two, repeated across trials,
so resampling whole tasks widens its interval to $[0.36,1.00]$. A verifier that approves everything scores
exactly $0$ while passing all 42 failing trajectories.

\begin{table}[t]
\centering\footnotesize
\begin{minipage}[t]{0.56\textwidth}
\centering
\setlength{\tabcolsep}{3.5pt}
\begin{tabular}{lrrrrr}
\toprule
Verifier $v$ & $\sigma(v)\uparrow$ & $\alpha\downarrow$ & $\varphi\downarrow$ & Admits & $n$ \\
\midrule
$\textsc{db}\times\textsc{comm}$ (ref.) & $1.00^{\dagger}$ & $0.00$ & $0.00$ & $0.48$ & 80 \\
\textsc{db} equality alone & $1.00^{\ddagger}$ & $0.00$ & $0.00$ & $0.48$ & 80 \\
LLM judge, NL assertions & $0.70$ & $0.10$ & $0.21$ & $0.43$ & 80 \\
Reference-action match & $0.67$ & $0.21$ & $0.12$ & $0.53$ & 72 \\
Communication substrings & \textsc{n/a} & --- & $0.00$ & $1.00$ & 4 \\
Constant (admit all) & $0.00$ & $1.00$ & $0.00$ & $1.00$ & 80 \\
\bottomrule
\end{tabular}
\caption{Verifier strength on the airline test split: one set of trajectories re-scored by each verifier
$v$; $n$ is the number of trajectories it applies to. $\alpha$ and $\varphi$ are false-admit and
false-reject rates against the reference ($\uparrow$/$\downarrow$: higher/lower is better). Errors cluster by task (App.~\ref{app:sigmamin}). $\dagger$~1.00 by
definition. $\ddagger$~1.00 because the communication checks pass on all 4 trajectories they apply to, so
\textsc{db} equality agrees with the reference on all 80. \textsc{n/a}: $\sigma$ is undefined because the
reference passes all 4 trajectories this check applies to, leaving nothing to correlate.}
\label{tab:sigma}
\end{minipage}\hfill
\begin{minipage}[t]{0.40\textwidth}
\centering
\setlength{\tabcolsep}{4pt}
\begin{tabular}{rrrr}
\toprule
$\varepsilon$ & $\sigma$ & Regret$\downarrow$ & Calib.\ gap$\downarrow$ \\
\midrule
$0.00$ & $1.000$ & $+0.0000$ & $+0.2000$ \\
$0.10$ & $0.802$ & $-0.0007$ & $+0.1604$ \\
$0.20$ & $0.602$ & $+0.0026$ & $+0.1311$ \\
$0.30$ & $0.399$ & $+0.0164$ & $+0.1255$ \\
$0.40$ & $0.199$ & $+0.0328$ & $+0.1308$ \\
$0.50$ & $0.001$ & $+0.0671$ & $+0.1573$ \\
\bottomrule
\end{tabular}
\caption{Experiment A2 (airline, eight candidates); $\sigma$ set by flipping reference verdicts with
probability $\varepsilon$. \emph{Regret}: held-out loss from picking with the corrupted verifier.
\emph{Calib.\ gap}: the picked candidate's reported training score minus its held-out score; 0 means the
loop reports its gain accurately. $\downarrow$: lower is better.}
\label{tab:arma}
\end{minipage}
\end{table}

\paragraph{Weaker verifiers pick worse changes, but not significantly (A2).} To get a fine-grained $\sigma$
scale cheaply, we flip each reference verdict with probability $\varepsilon$, which gives a verifier of known
strength; re-scoring one set of trajectories yields every point, each averaged over 400 random draws.
\emph{Regret}, how much worse the picked change is than the one a perfect verifier would pick, rises from
$0.000$ at $\sigma{=}1$ to $0.067$ at $\sigma{\approx}0$ (Table~\ref{tab:arma}), as predicted, but the fitted
slope ($-0.065$, CI $[-0.234,0.120]$) includes zero (one-sided $p{=}0.23$); resolving it would take about 220
held-out tasks, eleven times our 20. The bottleneck is the candidates, not $\sigma$: the
candidate a perfect verifier would pick is only $0.069$ above the average one, so even a random pick loses little, and the observed $0.067$
is already $98\%$ of that maximum (App.~\ref{app:pool}). Retail shows no trend (slope $+0.008$, CI
$[-0.115,0.109]$), but its candidates are so similar that replication is untested rather than failed.

\paragraph{Two secondary observations (A).} With a perfect verifier the loop still overstated its own gain
($0.20$; one-sided $p{=}0.08$, not in retail), and two verifiers with nearly equal $\sigma$ erred in opposite
directions (Table~\ref{tab:sigma}; App.~\ref{app:pool},~\ref{app:sigmamin}).

\paragraph{Only a stored copy restores behavior (B).} With the verifier fixed at the reference, a null
change sets the noise floors ($\rho_0{=}0.95$, CI $[0.89,1.00]$; $\beta_0{=}0.08$, CI $[0.02,0.16]$).
Restoring a prompt edit from a stored copy ($g{=}0$) returns behavior to the floor: $\rho-\rho_0{=}0.000$ (CI
$[-0.038,0.038]$), equivalent to zero within $\pm0.05$ (TOST $p{=}0.007$). Without a stored copy ($g{=}1$)
the undo must be regenerated, and three regenerated undos left behavior measurably different from the
baseline: $\rho-\rho_0{=}{-}0.16$ (CI $[-0.27,-0.07]$, one-sided $p{=}0.0001$; each undo significant after
Holm correction), measured on the training split against a floor of $0.98$. The gate refused all three, as the rule requires (App.~\ref{app:rev}).

\paragraph{Admitted changes reach tasks that rollback cannot undo (B).} Deployed on all 50 probe tasks, the three admitted changes with complete data altered
the outcome of $17\%$ of tasks beyond the noise floor
($\beta-\beta_0{=}0.17$, CI $[0.08,0.27]$, one-sided $p{=}0.0002$), and each change does so on its own after
Holm correction. Tasks flipped in both directions in nearly equal numbers (18 up, 20 down), so the average
effect is near zero while the reach is not, and reach is what rollback cannot
undo~\citep{garciamolina1987sagas}.

\begin{table}[t]
\centering\footnotesize
\setlength{\tabcolsep}{4pt}
\begin{tabular}{@{}lp{0.40\linewidth}p{0.42\linewidth}@{}}
\toprule
Precondition & Finding & Implication for unattended changes \\
\midrule
Verifier strength & Platform's LLM gate gave a constant in 9/37 settings, no error (\S\ref{sec:admission}) & A gate can look healthy while approving anything: measure $\sigma$ \\
Verifier strength & Weaker verifiers picked worse changes; not significant at 20 tasks (A2) & $\sigma_{\min}$ must be calibrated per platform (App.~\ref{app:sigmamin}) \\
Recoverability & Saved copy restores exactly; regenerated undo does not (B) & Only changes with $g{=}0$ can be undone without a human \\
Recoverability & A deployed change altered $17\%$ of tasks beyond noise (B) & Undoing a change does not repair what it already did \\
\bottomrule
\end{tabular}
\caption{Each result, the precondition it tests, and what it implies for unattended self-improvement.}
\label{tab:summary}
\end{table}

\section{Related work and discussion}

Production ML platforms made training and serving platform
services~\citep{baylor2017tfx,breck2019validation,moritz2018ray,sculley2015debt}, but not deciding what to
change in a deployed agent. Self-improvement and prompt-optimization
methods~\citep{shinn2023reflexion,madaan2023selfrefine,zelikman2022star,khattab2024dspy} supply the changes
we schedule but assume a trustworthy verifier; work on reward gaming and LLM judges questions that
assumption~\citep{skalse2022gaming,gao2023overopt,zheng2023judging}, and verifiable-reward methods avoid
it~\citep{lambert2024tulu,pezeshkpour2026autopyverifier}. Deployment engineering gates changes on
measured evidence~\citep{humble2010cd,beyer2016sre,k8sadmission} but assumes a human author. Closest is off-policy evaluation,
which promotes only when a lower confidence bound beats the
incumbent~\citep{thomas2015hcpi,thomas2016dope,sakhi2024logsmoothing}; $\sigma$ plays that role where no
unbiased value estimate exists.

\paragraph{Limitations.} The rule comes from one platform and was applied after the fact; only prompt edits were tested end to end, on
$\tau^2$-bench airline and retail rather than in production; and experiment A was too small to confirm the
$\sigma$ effect (App.~\ref{app:pool}).

\paragraph{Conclusion.} A self-improvement loop may run without a human only where the old version is kept and
the verifier's strength is measured; where either is missing, the platform should refuse.

\typeout{=== CONTENT ENDS ON PAGE \thepage\space ===}

\section*{Acknowledgments}
We thank Will Lu and Frank Petterson for their executive sponsorship of this work.

\appendix

\section{The gate case study in detail}
\label{app:design}

\paragraph{Setup.} A training set is built from a few \emph{seed tasks}; the number of distinct seed tasks is
its \emph{breadth}, and the share of examples coming from the most common seed task is its \emph{imbalance}.
We sent the production diversity gate, with its production prompt, training sets whose imbalance we set by
hand, holding everything else fixed (number and quality of examples, agent, task list). Each setting was run
10 times. The runs came in five batches (Table~\ref{tab:cells}): 172 settings and 1,720 calls, of which 1,718
returned a score (the two others answered in plain text when the scoring tool was offered but not required). The script and data are available
from the authors on request.

\begin{table}[htbp]
\centering\small
\caption{Scores from the diversity gate on the four-tool agent. Each column is an imbalance level: the share
of training examples that come from the most common seed task (the lowest possible share is .33 with 3 seed
tasks and .17 with 6). Each cell averages 10 runs. \textsc{no-signal} removes all imbalance information from
the prompt, so it cannot respond to imbalance; \textsc{threshold} states the rule as a number. $r_s$ is the
rank correlation between score and imbalance (--- when the score never changes).}
\label{tab:judge}
\begin{tabular}{llccccr}
\toprule
& & \multicolumn{4}{c}{largest-seed share} & \\
\cmidrule(lr){3-6}
& & .33/.17 & .50 & .67 & .83/.79 & $r_s$ \\
\midrule
breadth 3 & deployed          & 0.250 & 0.250 & 0.250 & 0.250 & --- \\
          & \textsc{no-signal} & 0.250 & 0.250 & 0.250 & 0.250 & --- \\
          & \textsc{threshold} & 0.250 & \textbf{0.100} & \textbf{0.100} & \textbf{0.090} & $-0.845$ \\
\midrule
breadth 6 & deployed          & 0.938 & 0.506 & 0.445 & 0.371 & $\mathbf{-0.836}$ \\
          & \textsc{no-signal} & 0.842 & 0.803 & 0.891 & 0.832 & $+0.063$ \\
          & \textsc{threshold} & 0.919 & \textbf{0.100} & \textbf{0.098} & \textbf{0.094} & $-0.786$ \\
\bottomrule
\end{tabular}
\end{table}

\begin{table}[htbp]
\centering\small
\caption{The five batches of runs. The judge models are \texttt{claude-haiku-4-5} (used in production),
\texttt{claude-sonnet-5} and \texttt{claude-opus-5}. The prompt variants are \textsc{shipped} (production),
\textsc{stat-only} (the imbalance instruction deleted, the number kept), \textsc{threshold} and
\textsc{no-signal}.}
\label{tab:cells}
\begin{tabular}{llrr}
\toprule
batch & what varies & settings & calls \\
\midrule
A & imbalance $\times$ judge model, breadth 3 and 6 & 24 & 240 \\
B & imbalance $\times$ prompt variant, breadth 3 and 6 & 32 & 320 \\
C & imbalance $\times$ how the score is requested & 24 & 238 \\
R & repeat of one dataset pair & 24 & 240 \\
S & imbalance $\times$ breadth $\in\{3,4,5,6,8,12\}$, two agents & 68 & 680 \\
\midrule
  & total & 172 & 1{,}718 \\
\bottomrule
\end{tabular}
\end{table}

\paragraph{The collapse is real.} The judge models do not let us set the sampling temperature, so every call
used the provider's default. Scores vary from run to run in 98 of the 172 settings, so a score that never
changes is not explained by deterministic decoding. We call a configuration \emph{collapsed} when it gives
the same score at all four imbalance levels. Nine of 37 configurations collapsed, all with the production
judge. That judge was also used in 33 of the 37 configurations, so this count on its own is not significant
(Fisher exact test, $p{=}0.55$). The stronger evidence is batch A, which gives all three judges identical
inputs: at breadth 3 the production judge returns the same score at every imbalance level, while both larger
judges change their score with imbalance ($p<10^{-11}$ each). Collapse also depends on the agent: the six-tool
agent tracks imbalance perfectly at breadth 3 but collapses at breadths 4, 5 and 8.

\paragraph{The fix works.} With the \textsc{threshold} wording, the gate scored at most 0.10 in all 16
settings where the most common seed task supplies more than half of the examples. On identical inputs this
happened in 30 of 30 runs, against 0 of 30 with the original wording (Fisher exact test, $p<10^{-16}$).

\paragraph{The second gate.} A second gate removes low-scoring rows from training data. It ships switched
off, because a safe threshold turned out to differ from agent to agent, and keeping unscored rows stops an
unreliable judge from quietly shrinking the training set.

\section{The two acceptance tests}
\label{app:tests}

\paragraph{Test 1: delete the signal.} Copy the gate's real input and remove the property it is supposed to
judge, entirely rather than setting it to a neutral value, so the prompt reads the same whatever the property's
value (the \textsc{no-signal} variant above). Run both versions the same number of times. If the scores are
the same, the gate is ignoring the property. Ours took 40 calls. Repeat the test whenever the model, the prompt
or the request format changes, because \S\ref{sec:admission} shows a gate can work in one setup and fail in
another.

\paragraph{Test 2: check the guarantees.} List the request settings the gate depends on, such as requiring a
tool call, requiring a fixed output format, or setting the temperature. For each one, send an input that
would be answered differently if the setting were ignored, and check that the answer changes. Three of ours
failed. Requiring a tool call cannot be combined with extended reasoning: the API rejects the request. Our
company's gateway forwarded requests for a fixed output format \emph{without} the format, and said nothing.
And the judge models reject temperature settings. We measured these on 2026-08-13 with one provider through
one gateway; both change over time, so each setting must be tested, not taken from documentation.

\section{Recoverability in detail}
\label{app:rev}

\begin{table}[htbp]
\centering\small
\caption{How each type of change is undone on our platform. All four keep the old version, so undoing is a
write and $g{=}0$ (for weights, switching back to the previous adapter). None of them can take back what the
change already did to users.}
\label{tab:rev}
\begin{tabular}{llll}
\toprule
dimension & how it is undone & $g$ & are its side effects undone? \\
\midrule
system prompt    & write back previous value & 0 & no: answers already served \\
tool definitions & write back previous value & 0 & no: calls already made \\
skills           & write back previous value & 0 & no: answers already served \\
model weights    & re-pin previous adapter    & 0 & no: answers already served \\
\bottomrule
\end{tabular}
\end{table}

\paragraph{Noise floors.} The agent and the simulated user are both stochastic, so running the \emph{same}
agent twice already changes some outcomes. We measure this by running the unchanged agent twice (the null
change). Its restoration fidelity is $\rho_0{=}0.95$ (CI $[0.89,1.00]$) on the 20 test tasks, and its blast
radius is $\beta_0{=}0.08$ (CI $[0.02,0.16]$) on all 50 airline tasks with four runs per task. We first used
20 tasks for blast radius, but the floor was too noisy, so we moved to 50; more tasks helped, more runs per
task did not.

\paragraph{Restoring a saved copy ($g{=}0$).} Writing back the saved prompt gives $\rho{=}0.95$, so
$\rho-\rho_0{=}0.000$ (CI $[-0.038,0.038]$). An equivalence test (TOST) with a margin of $\pm0.05$, one task in
20, confirms that the restored agent behaves like the original ($p{=}0.007$). Three separate runs of the
original agent agree with each other no more closely ($\rho$ of 0.950, 0.950 and 0.975). This is the expected
result, since the restored prompt is identical, but it is the control the other comparisons rely on.

\paragraph{Regenerating the undo ($g{=}1$).} Here we hid the old prompt, as when a fine-tuned model's previous
version is gone, and asked the optimizer to write an undo. It wrote three. We compared each with the original
agent on the 30 training tasks, against a floor measured with the same two runs per task ($\rho_0{=}0.98$).
The three undos reach $\rho$ of 0.75, 0.83 and 0.90, that is, 0.23, 0.15 and 0.08 below the floor
(one-sided $p{=}0.0005$, 0.004 and 0.04, all still significant after Holm's correction for testing three);
together, 0.16 below (CI $[0.07,0.27]$, $p{=}0.0001$). So a regenerated undo does not bring the original
behavior back, while a saved copy does; the two differ by about 0.16. Three caveats: these runs use the
training tasks, not the test tasks used for the saved copy; the gate refused all three undos, so this shows
what would have happened had one been installed; and the three are close rewrites of one another, so they
test one kind of undo rather than many. The gate refused them because their training pass rates (0.567,
0.517, 0.517) were below the original's (0.617), though that gap is too small to be significant ($-0.05$, CI
$[-0.23,0.13]$).

\paragraph{Blast radius.} We deployed four admitted changes, one at a time, in place of the original agent and
ran each on the 50 tasks. One change's test runs were lost, so we report the other three: blast radius 0.24,
0.28 and 0.24, against the 0.08 floor. Together they exceed the floor by 0.17 (CI $[0.08,0.27]$, one-sided
$p{=}0.0002$; a permutation test and a Wilcoxon test agree), and each change does on its own (exact McNemar
tests, $p{=}0.029$, 0.003 and 0.019; at most 0.04 after Holm's correction). Tasks flipped from pass to fail
about as often as from fail to pass (20 and 18), so the average pass rate barely moved even though about one
task in six changed outcome.

\section{Experiment A in detail}
\label{app:pool}

\paragraph{Why 20 test tasks are too few.} A task's outcome is close to all-or-nothing: its four runs usually
agree, and the same task often flips completely between one candidate and another. So differences between
candidates are mostly whole tasks, and with 20 test tasks the uncertainty on a candidate's test score is about
$\pm0.24$. Confirming the regret trend would take about 220 test tasks, and confirming the overstated gain
about 60.

\paragraph{Intervals.} All intervals come from resampling tasks with replacement: 2,000 resamples in
experiment A and 20,000 elsewhere.

\paragraph{How the regret slope is fit.} For each corruption level $\varepsilon$ from 0 to 0.5 in steps of
0.05 (Table~\ref{tab:arma} shows every other level), we measure $\sigma$ and the regret, each averaged over
400 random corruptions. The slope is a straight-line (least-squares) fit of regret against $\sigma$ through
these 11 points; a negative slope means regret grows as the verifier gets weaker. For its interval we resample
the training and test tasks 2,000 times and refit the whole curve each time; the one-sided $p$-value is the
share of refitted slopes that are zero or positive.

\paragraph{When can a candidate pool show the effect at all?} A verifier with $\sigma{=}0$ picks a candidate
at random, so the most it can lose is the gap between the average candidate and the one a perfect verifier
would pick. If that gap is smaller than the measurement noise, no experiment on this pool can show the effect;
we call such a pool \emph{field-limited}, and we fixed the threshold (gap divided by noise below 1) in advance.
Airline's gap is 0.069, a ratio of 0.62, and retail's is negative ($-0.009$), so both pools are field-limited.
Our first version of this check used a rougher estimate that put airline at 1.39; we corrected it after the
runs. We do not combine domains, because the pools with the least spread also have the smallest error bars
and would dominate a combined estimate.

\paragraph{Widening the pool.} The loop's own four revisions gave a ratio of at most 0.42, so we added
candidates made by deleting one section of the agent's instructions (no-cancel, no-book, no-modify, and two
pairs of these), each keeping at least 60\% of the instructions so that it is weakened rather than broken. One
candidate's test runs were lost, leaving eight, with test pass rates from 0.26 to 0.60.

\paragraph{The overstated gain.} The 0.20 gap at $\varepsilon{=}0$ has two parts: a difference between the
training and test tasks that affects every candidate (0.09, CI $[-0.09,0.26]$, one-sided $p{=}0.17$), and the
extra gain of the candidate that was picked because it scored best on training, known as the winner's curse
(0.11, CI $[-0.08,0.33]$, $p{=}0.13$). Neither part is significant, and neither is the total ($p{=}0.08$). In
retail the gap is $-0.02$ (CI $[-0.18,0.18]$), so it does not replicate.

\paragraph{Data checks.} A simulation that fails because of an infrastructure error still takes its place in
the results, with an empty transcript and no score, and the benchmark runner reports no error. We therefore
checked every set of runs before scoring and only compared runs with the same number of runs per task.

\section{Choosing $\sigma_{\min}$ and limiting false admits}
\label{app:sigmamin}

\paragraph{How to set $\sigma_{\min}$.} (i)~Decide how much regret $r$ the domain can tolerate, from the cost of
admitting a bad change. (ii)~Build a candidate pool that is not field-limited (App.~\ref{app:pool}); if that is
not possible, no threshold can be estimated. (iii)~Measure regret against $\sigma$ as in Table~\ref{tab:arma},
and set $\sigma_{\min}$ to the smallest $\sigma$ at which the \emph{upper} end of the confidence interval on
regret is at most $r$. (iv)~Repeat whenever the number of runs the gate averages over changes: the strength at
which regret starts to rise was 0.41 with 60 runs per candidate, 0.56 with 30 and 0.49 with 15.

\paragraph{Limit false admits, not only $\sigma$.} Let $\alpha$ be the false-admit rate (how often the
verifier passes a trajectory the reference fails) and $\varphi$ the false-reject rate. Our LLM judge has
$(\alpha,\varphi){=}(0.10,0.21)$ and reference-action match $(0.21,0.12)$: almost the same $\sigma$ but
opposite mistakes, although with 20 tasks the difference is not significant. Only false admits let a worse
change through, so where those are costly the rule should also require the upper confidence bound on $\alpha$
to stay below a limit $\alpha_{\max}$, with $\sigma>0$ to rule out collapse. Because mistakes repeat across a
task's runs, the bound must count tasks rather than runs: the judge's 4 false admits are one task seen four
times, which puts the upper bound near 0.37, not the 0.20 that counting runs suggests. Showing
$\alpha\le0.05$ with no false admits at all takes about 59 separate tasks that the reference fails.

\paragraph{Keeping $\sigma$ up to date.} $\sigma$ belongs to a verifier \emph{together with} its model and
settings, since the same gate collapsed in some setups and not others. A $\sigma$ measured before such a change
counts as unmeasured, so the rule refuses. The reference does not have to be a program, only fixed: label a
balanced set of trajectories once and re-score that same set each time. Where no program check exists, one can
be learned from labeled examples~\citep{pezeshkpour2026autopyverifier}, but it must then be measured on
examples it was not learned from.

\paragraph{Correcting the overstated gain.} The winner's-curse part can be removed by promoting a candidate on
the lower end of its confidence interval rather than on its best training
score~\citep{thomas2015hcpi,thomas2016dope,sakhi2024logsmoothing}, or reduced with statistical corrections for
selecting the best of several~\citep{andrews2024winners,efron2011tweedie,zollner2007winners}. The part that
affects every candidate needs a set of test tasks the gate never uses for selection. We recommend both; we
have not yet compared them on our data.

\end{document}